# Human-AI Collaboration or Academic Misconduct? Measuring AI Use in Student Writing Through Stylometric Evidence

**A PREPRINT**

Eduardo Araujo Oliveira[a,1], Madhavi Mohoni[a],

Sonsoles López-Pernas[b], Mohammed Saqr[b]

[a]School of Computing and Information Systems, University of Melbourne, Australia
[b]School of Computing, University of Eastern Finland, Finland

**ABSTRACT:** As human-AI collaboration becomes increasingly prevalent in educational contexts, understanding and measuring the extent and nature of such interactions pose significant challenges. This research investigates the use of authorship verification (AV) techniques not as a punitive measure, but as a means to quantify AI assistance in academic writing, with a focus on promoting transparency, interpretability, and student development. Building on prior work, we structured our investigation into three stages: dataset selection and expansion, AV method development, and systematic evaluation. Using three datasets — including a public dataset (PAN-14) and two from University of Melbourne students from various courses — we expanded the data to include LLM-generated texts, totalling 1,889 documents and 540 authorship problems from 506 students. We developed an adapted Feature Vector Difference AV methodology to construct robust academic writing profiles for students, designed to capture meaningful, individual characteristics of their writing. The method's effectiveness was evaluated across multiple scenarios, including distinguishing between student-authored and LLM-generated texts and testing resilience against LLMs' attempts to mimic student writing styles. Results demonstrate the enhanced AV classifier's ability to identify stylometric discrepancies and measure human-AI collaboration at word and sentence levels while providing educators with a transparent tool to support academic integrity investigations. This work advances AV technology, offering actionable insights into the dynamics of academic writing in an AI-driven era.

**Keywords:** Human-AI Collaboration, Authorship Verification, Artificial Intelligence in Education

## 1. Introduction

Human-Artificial Intelligence (HAI) collaboration in writing offers opportunities to enhance efficiency and boost student confidence; however, it also carries risks, such as reduced creativity, over-reliance on AI-generated content, and academic integrity (Kim & Lee, 2023). While the ethical use of AI in education is widely acknowledged as a way to enhance student learning (Cotton et al., 2023; Foltynek et al., 2023), the rise of Unauthorised Content Generation (UCG) presents a significant challenge to academic misconduct. Measuring the extent and nature of HAI collaboration in academic contexts remains a critical challenge for educators, particularly as generative AI (genAI) tools become increasingly available and integrated into educational settings (Atchley et al., 2024; E. Oliveira et al., 2023).

Distinguishing AI-generated text from human-authored content is necessary for understanding student learning behaviours, supporting skill development, and maintaining academic integrity. Analysing student writing patterns can help educators evaluate how students engage with AI tools, track their writing skill progression, and identify areas where additional support is needed (Pan et al., 2025). Existing detection tools for AI-assisted misconduct often lack reliability, explainability, and resilience to circumvention strategies such as paraphrasing (Cotton et al., 2023).
These challenges highlight the need for innovative, transparent, and robust approaches to address the unacknowledged use of genAI in HAI collaboration within academic writing (Kasneci et al., 2023).

---
[1] Corresponding author: eduardo.oliveira@unimelb.edu.au

Authorship Verification (AV)—a method traditionally used to determine whether two documents were authored by the same individual (Koppel & Schler, 2004)—offers a promising yet underexplored solution in educational contexts. In this study, we adapt and expand an AV technique to move beyond binary "yes" or "no" responses, instead leveraging the foundational principles of document comparison in a novel way. Based on student writing profiles, we propose a new approach that quantifies AI assistance in academic writing, focusing on stylistic granularity to enhance both transparency and interpretability. This adaptation aims to provide educators with actionable insights into the nature of HAI collaboration, supporting academic integrity while promoting responsible and informed use of AI in education.

## 2. Background Literature

### 2.1. The Role of AI in Education and Its Implications for Academic Integrity

The emergence of generative AI tools like ChatGPT offers transformative potential in education, enabling personalised learning, fostering critical discussions, and providing instant feedback (Cotton et al., 2023). However, these same capabilities risk undermining academic integrity by allowing students to delegate assignments to AI, raising concerns about the development of genuine writing and critical thinking skills.

Existing AI detection methods, such as OpenAI's GPT-2 detector (Kirchner et al., 2023), DetectGPT (Mitchell et al., 2023), and watermarking (Zhao et al., 2023), face challenges in educational contexts, including high false positive rates and difficulty analysing blended human-AI compositions (Weber-Wulff et al., 2023). These tools can unfairly penalise non-native English speakers, whose simpler writing styles are often misclassified as AI-generated (Kim & Lee, 2023). Addressing these limitations requires more equitable, transparent methods capable of capturing nuanced human-AI collaboration patterns.

Research into HAI collaboration in writing has highlighted its importance for understanding how AI influences skill development. For instance, (Gebreegziabher et al., 2023) found that higher-performing students actively refine AI-generated content, while lower-performing students rely passively on AI, limiting deeper engagement. Similarly, Stanford's CoAuthor project (Hai, 2023) illustrates productive collaboration practices, demonstrating how AI can enhance, rather than replace, student effort. These findings underscore the need to move beyond binary detection of AI authorship. By leveraging AV techniques, educators can better understand and support responsible AI use, fostering transparent and inclusive systems that reflect diverse writing styles and promote meaningful learning (Elkhatat et al., 2023).

### 2.2. Academic Writing Profiles for Authorship Verification

Stylometric analysis, which investigates the distinctive writing styles of individuals, offers a promising alternative to addressing the challenges previously discussed (Juola, 2008). As a quantitative approach to studying linguistic style, stylometry employs statistical and computational techniques to uncover unique patterns in an author's writing, such as their choice of words, sentence structures, punctuation habits, and other textual features (E. A. Oliveira et al., 2020; E. Oliveira & De Barba, 2022). The style of a text can be analysed through a diverse array of stylistic attributes, including lexical properties (e.g., differences in word usage, sentence length, or character-based metrics like vocabulary richness and word-length patterns), syntactic characteristics (e.g., function word usage, punctuation styles, and part-of-speech patterns), structural features (e.g., text layout, organisation, and formatting choices), content-specific indicators (e.g., word n-grams), and distinctive idiosyncratic markers (e.g., typographical errors, grammatical mistakes, or other unusual text features) (Abbasi & Chen, 2008; Zheng et al., 2006).

These stylistic markers are instrumental in AV, as they serve to identify an author's distinctive writing style and determine whether a specific text aligns with that style (Koppel & Schler, 2004; Potha & Stamatatos, 2014). In a typical AV task, a document with disputed authorship ($d_u$) is compared to a collection of texts with verified authorship ($D_{known}$), resulting in a classification of True (same author) or False (different author) (Potha & Stamatatos, 2014). However, AV is inherently complex, as an author's style may vary intentionally across different texts or evolve naturally over time (Koppel & Schler, 2004). Despite this variability, stylometry relies on the assumption that an individual's writing style exhibits a level of consistency that is identifiable and measurable (Laramée, 2018).

Approaches to AV generally fall into two paradigms: instance-based and profile-based. The instance-based paradigm approaches each text sample from an author as an independent entity, analysing them individually. Conversely, the profile-based paradigm consolidates all of an author's text samples into a unified, overarching profile for analysis. This profile is then compared to that of the questioned document using a dissimilarity metric (Potha & Stamatatos, 2014).

Given the dynamic nature of students' vocabularies and the evolution of their writing skills throughout higher education, the profile-based approach is particularly suited to educational contexts (E. Oliveira & De Barba, 2022). Despite the potential of stylometry in AV, there is a notable gap in research exploring the use of stylistic markers to develop and manage academic writing profiles for AV tasks, particularly in scenarios involving generative AI in educational settings.

## 3. Current Research

This exploratory study aims to investigate the potential of utilising students' academic writing profiles to inform AV methods in educational contexts, particularly for detecting and minimising issues related to unacknowledged use of genAI in human-AI collaboration. Developing effective academic writing profiles requires a nuanced understanding of stylometry, including the evolution of students' vocabularies over time.

To guide this investigation, we formulated the following research questions:
> (RQ1) Can we construct robust and dynamic academic writing profiles for students that effectively capture their unique writing characteristics for use in AV methods?
> (RQ2) How effectively can these academic writing profiles distinguish between human-authored and genAI-generated texts in AV tests when addressing similar academic tasks?
> (RQ3) How resilient are these student academic writing profiles in identifying human-AI collaboration when genAI is explicitly instructed to mimic students' writing styles?

## 4. Methodology

Our new investigation, which extends previous concise work presented in (Oliveira et al., 2024), was structured into three stages to systematically address our research questions:

**Stage 1: Dataset Selection and Expansion**: This stage involved selecting, pre-processing, and expanding datasets, which provided the foundational data needed to support all research questions.

**Stage 2: Adapted FVD Authorship Verification Method Development**: In this stage, we developed an adapted version of the Feature Vector Difference (FVD) AV method. Here, we expanded the existing FVD AV technique to move beyond binary "yes" or "no" responses, instead leveraging the foundational principles of document comparison in a novel way. Our new approach based on student writing profiles was designed and adapted to quantify AI assistance in academic writing, placing the focus on stylistic granularity to enhance both transparency and interpretability.

**Stage 3: Evaluation of the Adapted FVD AV Method Based on Research Questions**: In this stage, we systematically evaluated the adapted FVD AV method in alignment with each research question:

- Addressing RQ1: Can we construct robust and dynamic academic writing profiles for students that effectively capture their unique writing characteristics for use in AV methods? To address RQ1, we constructed and evaluated academic writing profiles based on genuine student submissions from the University of Melbourne datasets. These profiles were developed using the adapted FVD AV method, ensuring they captured the key individual characteristics of each student's writing. The aim was to establish that these profiles are not only consistent but also reliable for AV purposes in academic settings.

- Addressing RQ2: How effectively can these academic writing profiles distinguish between human-authored and genAI-generated texts in AV tests when addressing similar academic tasks? To answer RQ2, we evaluated the student writing profiles against genAI-generated texts. Here, genAI (GPT-4) was tasked with responding to the same academic prompts that students had answered. This allowed us to assess the effectiveness of student profiles in distinguishing between authentic student submissions and AI-generated texts when both address similar academic tasks.

- Addressing RQ3: How resilient are these student academic writing profiles in identifying human-AI collaboration when genAI is explicitly instructed to mimic students' writing styles? For RQ3, we evaluated the resilience of the student academic writing profiles in distinguishing genuine human-AI collaboration from deceptive mimicry by genAI.

This phase aimed to simulate scenarios where AI tools are used not only as collaborative aids but also to imitate students' unique writing characteristics. This evaluation assesses how well the adapted AV method could detect nuanced stylometric discrepancies and attribute authorship accurately, thus providing insights into the method's capability to capture the interplay between human input and AI-generated content.

**4.1. Stage 1: Dataset Selection and Expansion**

The first stage of the study focused on selecting, pre-processing and expanding datasets necessary to address our research questions in Stage 3. Three primary datasets (Table 1) were included in our investigations: (i) PAN-14 English Essays (Stamatatos et al., 2014), (ii) University General Essays 2019 (MGE-19) (Oliveira et al., 2020), and (iii) University Software Engineering (SWEN) Reports 2021 (MSR-21) (Rios et al., 2023). PAN-14 dataset is well-suited to validate our adapted FVD AV method because it is a publicly available dataset collected in an academic environment and serves as a standard benchmark for testing AV methods. The public nature of PAN-14 enables direct comparison of our results with numerous other AV methods reported in the scientific community, facilitating meaningful evaluation and validation of our approach (Stamatatos et al., 2014). University General Essays 2019 (MGE-19) and University SWEN Reports 2021 (MSR-21) are real datasets from University of Melbourne: University General Essays 2019 (MGE-19) is a university-wide dataset, University SWEN Reports 2021 (MSR-21) is a domain-specific dataset.

*4.1.1. PAN-14 dataset*

The PAN-14 AV dataset originates from the Uppsala Student English (USE) corpus, comprising 1,489 English essays written between 1999 and 2001 by 440 Swedish students learning English as a Secondary Language (ESL) across three academic levels (Axelsson, 2002). The essays collectively contain 1,221,265 words, with an average length of 820 words per essay. Essays from the first term typically are shorter, with an average of 777 words. These essays address specific predetermined topics and vary in type. They were composed outside of class, with a submission deadline of two to three weeks, a word limit (generally between 700 to 800 words), and recommendations for appropriate text structure. The content of the essays encompasses discussions of literature, arguments, reflections, and personal narratives.

For the PAN 2014 competition AV task (Stamatatos et al., 2014), the USE corpus was filtered to essays with a minimum of 500 words per document, and 1–5 known essays per author, resulting in a dataset of 435 authors.

*Table 1.* Statistics of the dataset used in our research

| Dataset | Train size | Test size | Num. problems | Num. docs | Avg. known docs per problem | Avg. words per doc |
| --- | --- | --- | --- | --- | --- | --- |
| PAN-14 | 200 | 200 | 400 | 1447 | 2.6 | 840.6 |
| MGE-19 | 84 | 20 | 104 | 383 | 6.9 | 106.0 |
| MSR-21 | 18 | 18 | 36 | 59 | 2.8 | 1007.0 |

*4.1.2. University General Essays 2019 (MGE-19) Dataset*

The MGE-19 dataset (Ethics approval #1748727.1) was collected from university students between 2017 and 2019. 46 students from four main disciplines – Engineering (24%, n=11), Commerce (24%, n=11), Arts (19.5%, n=9), Science (13%, n=6) and other (19.5%, n=9) were tasked with six questions (Table 2) each corresponding to a different cognitive load (knowledge, comprehension, application, analysis, synthesis, evaluation) (Oliveira et al., 2020). Participants had 20 minutes to answer four open-ended questions requiring low to medium cognitive load (Q1, Q2, Q3, Q4). After the initial 4 questions, participants were given 30 minutes to answer two open-ended questions requiring medium to high cognitive load (Q5, Q6).

*Table 2.* MGE-19 survey questions

| ID | Question |
| --- | --- |
| Q1 | What made you decide to join this university? |
| Q2 | What would you say has been the best class you have taken at this university? What did you enjoy about that class? |
| Q3 | You are asked to complete a group assignment. It is important all students in the group contribute equally to the project. Come up with a plan for completing the group assignment, from research to class presentation. |
| Q4 | Describe the similarities and differences between preparing a written assignment and preparing for a final exam. |
| Q5 | A fellow university student spends a significant amount of their time worrying about their ability to complete their academic work, and becomes very concerned when they do not meet their grade expectations. In addition, they are concerned about financial pressures such as rent and textbook costs. Considering the texts you have received and the situation presented above, please answer the following question: Do you think the university should support this student to improve their well-being? Why or why not? |
| Q6 | [Using the scenario from Q5] Describe what advice you would provide to the student to help improve their well-being. What steps could they take? |

After obtaining the answers from participants, the dataset was examined and cleaned. Among all received answers, 21 had less than 25 words or 140 characters. Previous research has shown that significantly small text samples can impact the performance of AV (Stein et al., 2007), despite some examples of the contrary (Escalante et al., 2011). With this in mind, we excluded all texts with less than 25 words (which is approximately 140 characters), following the same approach as reported in (Oliveira & de Barba, 2022). We followed the PAN-14 framework to transform the texts to an AV problem set. Same-author instances were created by shuffling each author's texts and randomly removing one as the unknown text, while different-author instances were made by shuffling the known texts and randomly selecting a different author text as the unknown. This resulted in an equal number of positive and negative cases.

### *4.1.3. University SWEN Reports 2021 (MSR-21) Dataset*

The MSR-21 dataset (Ethics approval #24272) comprises written submissions from 20 university students enrolled in a yearlong software engineering project subject as part of their Master's in Software Engineering program. The subject includes 10 assessments spread over two teaching semesters: 6 team-based (group) assessments and 4 individual assessments. The individual assessments comprised a personal objectives statement, an analysis addressing a digital ethics issue, a reflective report on the subject experience, and a statement outlining individual contributions to their projects (Rios et al., 2023). The personal objectives task was carried out during a supervised classroom session and submitted through the Canvas Learning Management System (LMS), whereas the other assessments were completed independently and asynchronously.

After obtaining the answers from software engineering students, the dataset was examined and cleaned. Some students submitted assessments in different file formats: .pdf, .txt, .html, .docx. In some cases, we had to remove HTML tags, metadata, or special characters from these files. Similarly to the MGE-19 dataset, all texts with less than 25 words were excluded, and then the data was transformed into an AV problem set following the PAN-14 framework.

For all datasets considered in this study, we did not strip punctuation marks or numbers to students' texts to preserve specific individual characteristics in the writings.

*4.1.4. MGE-19-GPT and MSR-21-GPT Datasets Expansion*

Our aim for these expansions was to replicate the scenario of a disputed academic submission being authored by genAI, and to test the performance of our AV approach, trained only on human submissions. To create these expansions, we prompted GPT-4 to answer the same questions provided to students in the original MGE-19 and MSR-21 data collection tasks using the prompt shown in Table 3. We then replaced the unknown texts in the different-author problems of the test set with a randomly selected and identically pre-processed GPT-authored text.

For each AV test set of *m* different-author problems, we used OpenAI ChatCompletion API (GPT-4) to batch generate *m* responses ('choices') answering each of the questions provided to students. The system prompt outlined in Table 3 was designed to replicate realistic ChatGPT usage, hence we did not provide specific instruction on written tone or style, and opted for default parameters of *1* for *temperature* and *top_p*. We provided all questions in a single user prompt to GPT, and instructed the LLM to separate its answers with a specified delimiter (%%%), which we used to split different answers to each question later on. Aside from saving cost, this gave us the additional benefit of a similar average word length for MGE-19-GPT (97.76 words) compared to MGE-19 (106.0 words). We found that phrasing the prompt to "*Answer the following essay questions from an <educational context> survey*" instead of "*Pretend you are a <educational context> student*" circumvented the behaviour of GPT to sometimes include disclaimers that it is an AI chatbot at the end of its answers. Further instruction was provided to prevent the inclusion of question numbers for uniformity and to avoid the need for extra post-processing.

For MSR-21-GPT, given the larger average document lengths with fewer questions and authors, we placed one request for each task (providing the same system prompt followed by the task question). We had to provide more specification for the system prompt ("Answer from the perspective of a student") to circumvent the behaviour of GPT misinterpreting the instruction and responding with a rephrasing of the task itself.

*Table 3.* System prompts to provide context used for MGE-19-GPT and MSR-21-GPT

| MGE-19-GPT | MSR-21-GPT |
| --- | --- |
| Answer the following essay questions from the University of Melbourne Master's student survey. Write only the answers, without putting the questions or any answer numbers. Between every two answers put 3 percent signs like so: %%% | You are answering the following report essay tasks for students undertaking a yearlong team-based {subject code} Software Engineering Project, part of the Master's of Software Engineering degree. Answer from the perspective of a student. These tasks were designed for reflection at various points in the academic year. Give just answers without using questions, numbers or any other details. |

*4.1.5. MGE-19 and MSR-21 GPT Dataset Impersonation (GPT-I)*

The objective of this new data expansion process was to identify the extent to which our adapted AV method would be able to identify a GPT co-authored text, or one that has been obfuscated to appear as the original author. We provided GPT with text samples of an author (student) and a question to be answered in a similar writing style, asking AI to explicitly mimic authors' writing styles in generated answers. The system prompt contained impersonation instructions, the task prompt provided instructions to GPT (similar to the ones performed by students) and, the user prompt contained an example of text from that author (Table 4).

More specifically, if an author has texts $\{a_1,...,a_n\}$ which are answers to the task questions $\{1,...,n\}$, we select a particular sample answer $a_i$ to question *i*. We then select a different question *j* such that $j \neq i$ and provide $a_i$ and *j* to GPT to answer the question $a_j$ impersonating the style in $a_i$. To make an AV problem, we remove $a_j$ from the author's texts and set them as the known texts, and set the GPT impersonation response as the unknown text.

The dataset expansion was done in a similar manner to the GPT expansions described in the previous section . For every negative case (different-author) known and unknown texts, we replaced the unknown text with a GPT impersonation of the author of the known texts.

*Table 4.* System, task, and user prompts used for MGE-19-GPT-I and MSR-21-GPT-I

| | |
|---|---|
| **System** | Based on the following sample text and task provided, imitate the writing style identified in the sample text, including all idiosyncrasies, turns of phrase, parts of speech usage, vocabulary, structure, quirks, and writing habits presented in that. Try to produce a response for the provided task that could believably have been written by the same author of the sample, and within the context of the provided task. |
| **Task (or question)** | Instruction Provided to Student: {sample task} |
| **User** | Sample Text Question: {sample text} |

## 4.2. Stage 2: Adapted FVD Authorship Verification Method Development

After preprocessing and expanding the datasets, we proceeded with developing our AV method. This approach builds on the existing FVD method introduced by (Weerasinghe & Greenstadt, 2020) in their PAN-20 (Kestemont et al., 2020) submission. FVD operates by extracting a combination of common character n-grams (CNG) and handcrafted features, calculating the difference between two text vectors, and applying logistic regression to classify the authorship as either *True* (same author) or *False* (different author). Despite its straightforward CNG design, the FVD method demonstrated exceptional performance in PAN-20 (Kestemont et al., 2020), which involved fan-fiction texts with a closed set of authors, and in PAN-21 (Kestemont et al., 2021), which tested its robustness with unseen authors.

The method was selected for this study due to its effectiveness in utilising character n-grams, parts-of-speech (POS) n-grams, and custom features like vocabulary richness. The integration of these features with logistic regression provides not only reliable classification results but also an interpretable framework for understanding the underlying factors influencing the model's decisions.

Adaptations were performed to the traditional FVD method in our research to meet several key software requirements essential for future practical application of this method in educational settings:
- Performance: The adapted method should perform adequately on the standard AV performance metrics ($F_1$, $F_{0.5}$, $c@1$, *AUC*, *Brier Score*). Ensuring high performance across these metrics is important for the method's reliability and accuracy in detecting authorship discrepancies.
- Simplicity and explainability: We aimed to keep the FVD method both simple and explainable. This allows us to analyse the impact of different stylometric features on classification outcomes. In academic settings, relying on an uninformative "black box" classifier, even if highly accurate, can be detrimental. It is far more beneficial to provide transparent and explainable feature contributions, enabling assessors to make informed decisions using their judgment. To achieve this, we avoided overly abstract features and AV methods that offer limited insight into the classification process.
- Robustness to cross-topic settings and ability to handle shorter texts and small datasets: Our method needed to be robust across variations in topic and text length, reflecting the real-world academic environment. Unlike some AV tasks that must handle differences in genre, discourse type, language, and text length, our focus was on ensuring robustness to variations in topic and accommodating shorter texts and small datasets typical in educational contexts.

There are three main steps to our adapted FVD method: (1) data pre-processing, (2) feature extraction, and (3) classification. These are depicted in Figure 1. The following sections provide an overview of each of these steps.

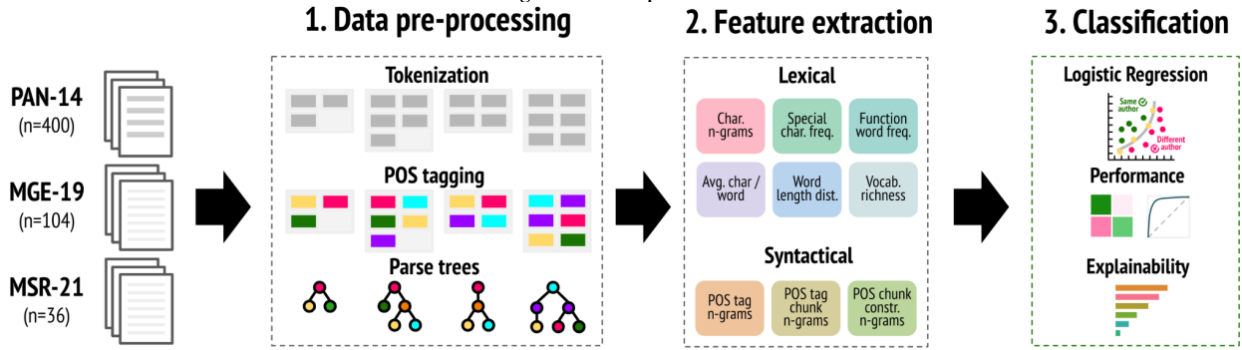
*Figure 1.* Adapted FVD

### 4.2.1. FVD Pre-processing

In the preprocessing step, each student text was prepared by breaking it down into tokens, identifying POS, and generating higher-level structures called parse method trees. These processes helped us to capture both the basic elements of language (words, punctuation) and their grammatical functions, forming a foundation for the feature extraction.

### 4.2.2. FVD Feature Extraction

After preprocessing, we extracted features from each text to create student writing profiles. These features were designed to capture different aspects of each student's writing style and are summarised in Table 5.
Key features included: (i) Lexical Features - These included character patterns (such as groups of three to six characters), function word frequencies, and vocabulary richness. The aim was to capture specific patterns that reflect an individual's writing style; (ii) Syntactic Features - These involved patterns of parts of speech and their combinations, as well as structures like noun and verb phrases. These features gave us insight into how students build sentences, revealing deeper stylistic elements of their writing. These features, used to build writing profiles, served as the basis for comparison in our AV process.

### 4.2.3. FVD Classification

We used logistic regression as our classifier, primarily because of its simplicity and explainability —important factors in an educational context. Logistic regression provides transparency by allowing us to see how each feature influences the final classification decision. While more complex classifiers could be used, such as Neural Networks, logistic regression's interpretability was crucial for explaining results to educators and understanding why certain texts were classified in a particular way.

A key part of our approach was explainability. For each trained model, we analysed features such as character patterns, parts of speech, and function word distributions. This analysis allowed us to gain insight into which aspects of students' writing were most influential in verifying authorship, providing transparency and interpretability, which are critical for educational assessments.

Table 5. Features extracted from text in the FVD method.

| Category | Feature | Description |
|---|---|---|
| **Lexical** | Character n-grams | TF-IDF values for character n-grams with 3 ≤ n ≤ 6 |
| | Special character frequencies | TF-IDF values of the following special characters: !"#$%&'()*+,-./:;<=>?@[\^_`{|}~ |
| | Function word frequencies | Using 179 stopwords available in the NLTK corpus |
| | Average characters per word | Average number of characters in a token |
| | Word length distributions (1-10) | Fraction of tokens of length *l*, where 1 ≤ *l* ≤ 10 |
| | Vocabulary richness | The ratio of hapax-legomenon (number of words appearing once) to dis-legomenon (number of words appearing twice) divided by the total number of tokens in the text (for scaling) |
| **Syntactic** | POS Tag n-grams | TF-IDF values of POS tag tri-grams. In the previous example, this would consider tri-grams over tags ['IN', 'PRP', 'VBP', 'TO', 'VB', 'DT', 'NN', 'NN', 'IN', 'NN', ',', ...] |
| | POS Tag chunk n-grams | TF-IDF values for POS tag chunk tri-grams (higher level of parse tree). For the sentence above, tri-grams are taken over chunks ['IN', 'NP', 'VP', 'NP', 'IN', 'NP', ',', 'NP', ...] |
| | POS chunk construction (NP, VP) n-grams | TF-IDF values of each noun phrase and verb phrase expansion. For the sentence above, tri-grams are taken over expansions ['NP[PRP]', 'VP[VB TO VB]', 'NP[DT NN NN]', 'NP[NN]', 'NP[PRP]', ...] |

### 4.3. Stage 3: Evaluation of the Adapted FVD AV Method Based on Research Questions

Following the development of our adapted AV method, we conducted a series of focused analyses to systematically evaluate its performance in relation to each of our research questions.

Addressing RQ1: Can we construct robust and dynamic academic writing profiles for students that effectively capture their unique writing characteristics for use in AV methods? To answer RQ1, we validated our adapted AV method, which makes use of academic writing profiles, using the PAN-14 dataset. This served as a benchmark to determine whether the adapted FVD method could perform at a level comparable to other leading CNG-based AV approaches. We employed the same evaluation framework used in PAN-14, utilising standard AV metrics such as c@1, AUROC, and the 'final score' (Stamatatos et al., 2014). Our results were benchmarked against the top-performing CNG-based approaches in PAN-14, including those documented by (Frery et al., 2014; Moreau et al., 2014; Satyam et al., 2014), as well as the PAN-14 baseline. By validating the method on the PAN-14 dataset, we affirmed its applicability to our institutional datasets (MGE-19 and MSR-21).

Addressing RQ2: How effectively can these academic writing profiles distinguish between human-authored and genAI-generated texts in AV tests when addressing similar academic tasks? For RQ2, we aimed to evaluate how effectively student academic writing profiles could perform in AV tests, particularly when compared to texts generated by genAI. To do so, we applied the adapted FVD method to the genuine student datasets (MGE-19 and MSR-21) and their corresponding MGE-19-GPT and MSR-21-GPT expanded datasets, where generative AI was tasked with completing similar academic tasks.

We used metrics like c@1, AUROC, and the 'final score,' along with ROC curve analysis, to assess our classifier's ability to distinguish genuine student texts from AI-generated ones. ROC curves evaluated performance across thresholds, while True Negative Rate (TNR) visualisations highlighted its effectiveness in rejecting false positives—critical for maintaining academic integrity. Additionally, a focused TNR analysis of altered negative cases from our expansion experiments enabled precise comparisons between genuine-author, different-author, and AI-generated test sets.

Addressing RQ3: How resilient are these student academic writing profiles in identifying human-AI collaboration when genAI is explicitly instructed to mimic students' writing styles? For RQ3, we sought to determine the resilience of the adapted FVD method when generative AI was explicitly instructed to mimic student writing styles. To answer this question, we applied our AV method to the MGE-19-GPT-I and MSR-21-GPT-I datasets, which included genAI-generated texts attempting to closely imitate the writing styles of specific students.

We applied a similar evaluation framework as the one adopted in RQ2. In this stage, a key focus was the classifier's robustness: whether it could still correctly attribute authorship despite deliberate attempts by genAI to mimic individual writing styles. Again, we specifically analysed TNR to determine the method's ability to correctly reject falsely generated mimicry attempts, ensuring our classifier was not easily fooled by genAI impersonation attempts.

## 5. Results

### 5.1. Addressing RQ1: Can we construct robust and dynamic academic writing profiles for students that effectively capture their unique writing characteristics for use in AV methods?

Table 6 provides a performance comparison between our adapted AV method, the top three CNG approaches from PAN-14 for the English Essays task, and the PAN-14 baseline. Aside from evaluation metrics, following the work of (Jankowska, 2017), the table also shows the number of PAN-14 submissions outperformed by our method in that dataset (out of 13 submissions).

*Table 6.* A comparative analysis of the performance of our adapted FVD and CNG methods relative to the top three methods and the baseline in the PAN-14 English Essays task (Stamatatos et al., 2014)

| Method | # methods outperformed | Final score | AUC | c@1 | # unanswered |
|---|---|---|---|---|---|
| (Frery et al., 2014) | 13 | 0.513 | 0.723 | 0.710 | 15 |
| (Satyam et al., 2014) | 12 | 0.459 | 0.699 | 0.657 | 2 |
| Adapted FVD | 12 | 0.430 | 0.680 | 0.633 | 6 |
| (Moreau et al., 2014) | 11 | 0.372 | 0.620 | 0.600 | 0 |
| PAN-14 baseline | 2 | 0.288 | 0.520 | 0.548 | 0 |

The PAN-14 English Essays task is particularly regarded as challenging (Jankowska, 2017), with most participants — including the overall PAN-14 winner (Khonji & Iraqi, 2014) — exhibiting significantly lower performance on this dataset relative to other PAN14 AV tasks. Nonetheless, our adapted method demonstrated relatively robust performance, achieving an AUC of 0.680 and a c@1 score of 0.633.

Our adapted FVD AV method exhibited a substantial improvement in explainability compared to earlier machine learning models. This enhanced interpretability is highlighted by the analysis of the most impactful features across all datasets (as shown in Figure 2). For the PAN-14 dataset, the adapted FVD method attributed the highest coefficient values to POS tag features during training. Among these, ('NNS', '.'), denoting a plural noun followed by punctuation, was the most prominent, often corresponding to sentences concluding with a plural noun. Additionally, character n-grams such as 'nur' and 'acters' emerged as key features. Informal analysis suggested these patterns were related to terms like 'nurturing' and 'nursing', and to a literature report centred on a poem featuring a prominent nurse character, potentially explaining these observed patterns.
Another notable POS chunk feature was ('NP', '.', ' " ')—a noun phrase followed by punctuation and a quotation mark—commonly occurring at the end of quoted text. High coefficients were also observed for noun phrases structured as NP[CD NNP], which indicate a cardinal number followed by a singular proper noun. This pattern appeared in examples like dates (e.g., "four March"), combinations of year and location (e.g., "1994 Sweden"), and occasionally mislabelled book titles (e.g., "One Flew Over the Cuckoo's Nest"). The function word 'haven' (as in 'haven't') was identified as the most significant among function words, though both function words and special characters generally contributed less weight to the model overall. By highlighting these patterns through user-friendly interfaces, this approach can enhance explainability in AV use in educational settings. These insights not only enable a nuanced understanding of students' writing styles but also foster constructive dialogue around academic integrity and writing development.

*Figure 2*. Example of analysis of academic writing - top five most influential features identified by our adapted FVD method during classification on the PAN-14 dataset

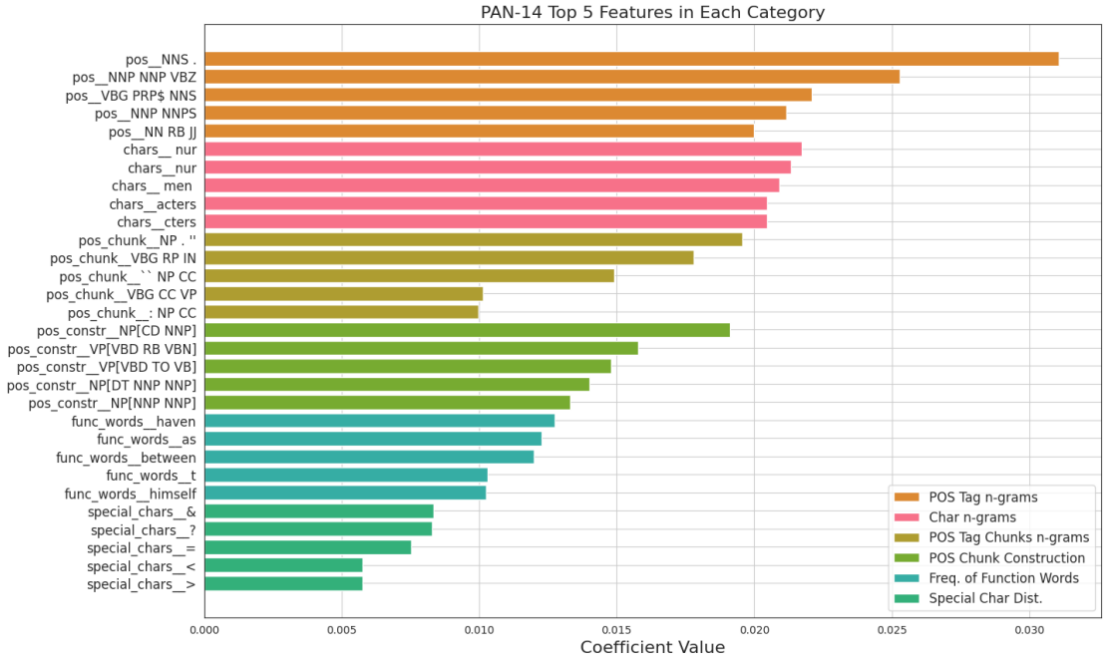

### 5.1.1. MGE-19 and MSR-21 datasets

The performance of FVD on MGE-19 and MSR-21 are shown in Table 7. FVD achieves an AUC of 0.791 on the MGE-19 dataset, and performs better on all metrics than our experiments on PAN-14. This improvement in performance is also reflected in the ROC as FVD is able to stay consistently higher than AUC = 0.5, demonstrating its ability to distinguish between positive and negative classes for this dataset (Figure 3). It also shows a moderately sharp initial uptick of TPR even at a lower decision threshold, which indicates better identification of true positives and conversely reduction of false negatives (false classification as different authorship). Although the dataset is balanced (equal positive and negative cases), FVD learns to classify nearly double the instances as close to $p = 1$ (same authorship) than close to $p = 0$ (different authorship).

The FVD method achieved an overall score of 0.761 on MSR-21, surpassing the results obtained on the previous two datasets, with a notably higher AUC of 0.828. The Average ROC curve consistently stays above the AUC = 0.5 line across all thresholds (Figure 5), indicating strong performance. While a stair-like pattern is visible, likely due to the limited size of the dataset despite testing across multiple folds, the sharp initial rise in the curve demonstrates the method's ability to accurately identify same-author cases even at lower decision thresholds.

*Table 7*. Evaluation of the Adapted FVD AV Method on the MGE-19 and MSR-21 Datasets.

|        | AUC   | c@1   | $F_{0.5}$ | $F_1$ | Brier | Overall |
|--------|-------|-------|-------|-------|-------|---------|
| MGE-19 | 0.791 | 0.726 | 0.697 | 0.770 | 0.780 | 0.753   |
| MSR-21 | 0.828 | 0.723 | 0.712 | 0.749 | 0.790 | 0.761   |

*Figure 3.* ROC curve comparison for MGE-19 and MGE-19-GPT datasets

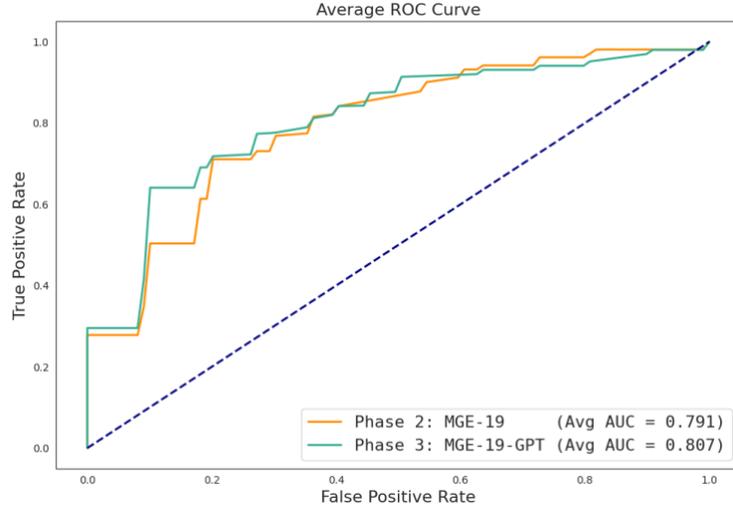

**5.2. Addressing RQ2: How effectively can these academic writing profiles distinguish between human-authored and genAI-generated texts in AV tests when addressing similar academic tasks?**

*5.2.1. MGE-19-GPT dataset*

MGE-19-GPT has a total of 104 modified test cases with GPT-authored texts as unknowns (out of 208 total test instances) after 5 folds over 2 repetitions with test sizes of 20-21 in each run). FVD's performance does not drop on MGE-19-GPT and in fact achieves a higher AUC (0.807) than MGE-19. However, its performance on other metrics is lower than on MGE-19, resulting in a slightly lower overall score of 0.749 (Table 8).

The ROC curves of the distributions are shown in Figure 3, with similar distances from AUC = 0.5 achieved by both. Given that only the negative cases in the test set have been modified, only a small change in ROC curve is to be expected unless the model experiences a significant failure on the negative cases.

The TNR comparison with MGE-19 is shown in Figure 4. FVD achieves a TNR of 60.7%, i.e. it correctly classified 60.7% of the GPT-authored texts as different authorship across all the folds of the experiment. This is 1.8% lower than the TNR achieved on the original dataset.

*Table 8.* Evaluation of the Adapted FVD AV Method on the MGE-19-GPT Dataset.

|  | AUC | c@1 | $F_{0.5}$ | $F_1$ | Brier | Overall |
|---|---|---|---|---|---|---|
| MGE-19-GPT | **0.807** | 0.715 | 0.691 | 0.766 | 0.770 | 0.749 |
| MGE-19 | 0.791 | **0.726** | **0.697** | **0.770** | **0.780** | **0.753** |

*Figure 4.* TNR comparison for MGE-19 and MGE-19-GPT datasets

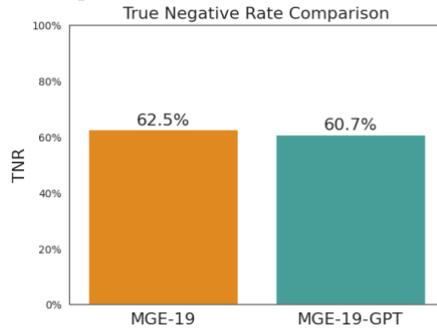

### 5.2.2. MSR-21-GPT dataset

Performance metrics reveal that MSR-21-GPT outperforms MSR-21 across all k folds. Of the 180 test cases in the dataset, 90 were modified, with evaluations conducted over 2 folds, 5 repetitions, and a test size of 18 per iteration. Table 9 showcases the FVD method's superior results on MSR-21-GPT, and Figure 5 underscores this with a consistently higher ROC curve at every threshold compared to MSR-21.

*Table 9.* Evaluation of the Adapted FVD AV Method on the MSR-21-GPT Dataset.

|  | AUC | c@1 | $F_{0.5}$ | $F_1$ | Brier | Overall |
|---|---|---|---|---|---|---|
| MSR-21-GPT | **0.875** | **0.772** | **0.767** | **0.787** | **0.839** | **0.808** |
| MSR-21 | 0.828 | 0.723 | 0.712 | 0.749 | 0.790 | 0.761 |

*Figure 5.* ROC curve comparison for MSR-21 and MSR-21-GPT

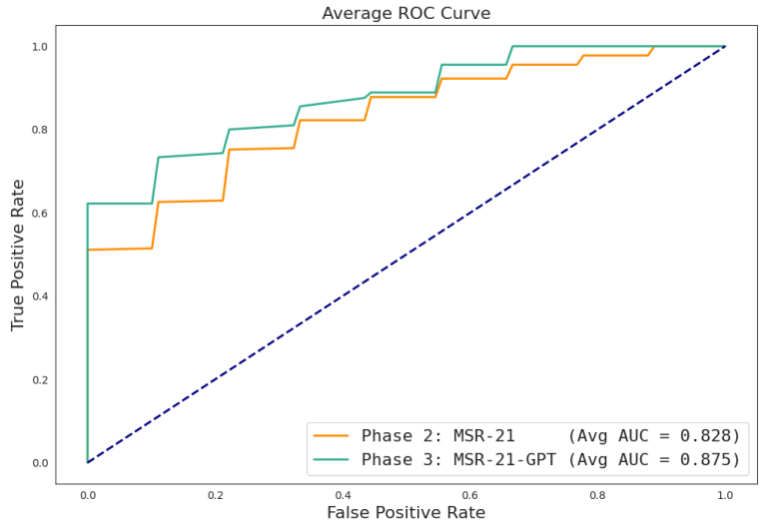

The TNR comparison is shown in Figure 6. FVD on MSR-21-GPT achieves an average TNR of 78.9% over all folds, 10% higher than MSR-21, indicating it was more able to correctly predict GPT-authored texts. Overall, the FVD method demonstrated superior performance on MSR-21-GPT compared to MSR-21 across all conducted comparisons

*Figure 6.* TNR comparison for MSR-21 and MSR-21-GPT datasets

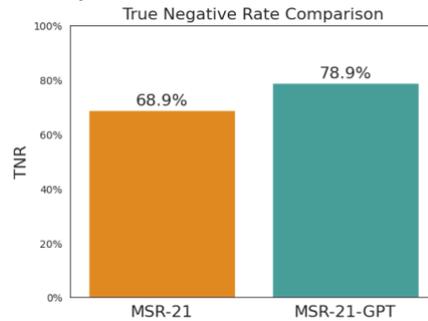

**5.3. Addressing RQ3: How resilient are these student academic writing profiles in identifying human-AI collaboration when genAI is explicitly instructed to mimic students' writing styles?**

*5.3.1. MGE-19-GPT-I dataset*

The MGE-19-GPT-I expansion has the same number of modified cases as MGE-19GPT (104 modified different author instances out of 208 total test instances). A greater performance on all evaluation metrics was achieved on MGE-19-GPT-I as shown in Table 10. The ROC curve shown in Figure 7 indicates an improved ROC at all thresholds compared to MGE-19. The TNR comparison in Figure 8 shows a 73.8% rate of correctly predicted different-author instances in MGE-19-GPT-I, an 11.3% increase from MGE-19. Overall, FVD performance on MGE-19-GPT-I is higher than performance on MGE-19 across all metrics considered.

*Table 10.* Evaluation of the Adapted FVD AV Method on the MGE-19-GPT-I Dataset.

|  | AUC | c@1 | $F_{0.5}$ | $F_1$ | Brier | Overall |
|---|---|---|---|---|---|---|
| MGE-19-GPT-I | **0.876** | **0.773** | **0.751** | **0.816** | **0.836** | **0.810** |
| MGE-19 | 0.791 | 0.726 | 0.697 | 0.770 | 0.780 | 0.753 |

*Figure 7.* ROC curve comparison for MGE-19 and MGE-19-GPT-I datasets

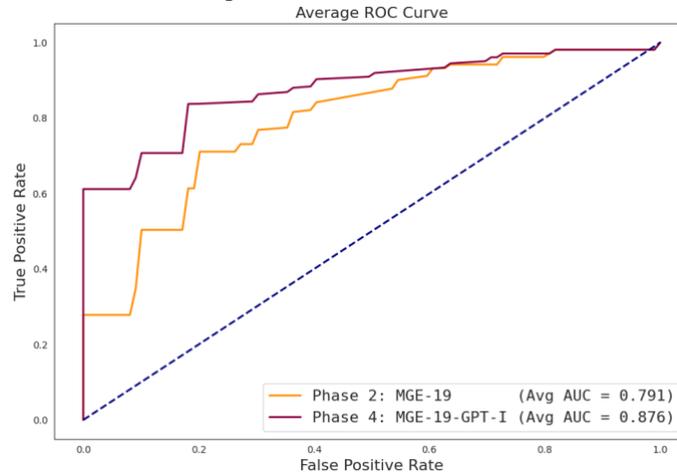

*Figure 8.* TNR comparison for MGE-19 and MGE-19-GPT-I datasets

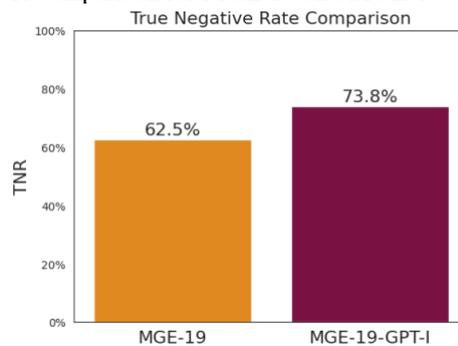

*5.3.2. MSR-21-GPT-I dataset*

MSR-21-GPT-I follows the same structure as MSR-21-GPT, comprising 180 test cases, half of which (90) are modified. Evaluated over two folds, five repetitions, and 18 cases per run, the performance metrics in Table 11 reveal that the expanded dataset outperformed MSR-21 on all measures.

The ROC curve of MSR-21-GPT-I shows better distinction between classes at all thresholds than MSR-21 (Figure 9). The TNR comparison in Figure 10 shows that 80% GPT impersonations were correctly classified as different authorship, 11.1% more than MSR-21 classification. Overall, FVD performs better on MSR-21-GPT-I than on MSR-21 across all metrics.

*Table 11.* Evaluation of the Adapted FVD AV Method on the MSR-21-GPT-I Dataset.

|  | AUC | c@1 | $F_{0.5}$ | $F_1$ | Brier | Overall |
|---|---|---|---|---|---|---|
| MSR-21-GPT-I | **0.889** | **0.787** | **0.779** | **0.789** | **0.841** | **0.817** |
| MSR-21 | 0.828 | 0.723 | 0.712 | 0.749 | 0.790 | 0.761 |

*Figure 9.* ROC curve comparison for MSR-21 and MSR-21-GPT-I datasets

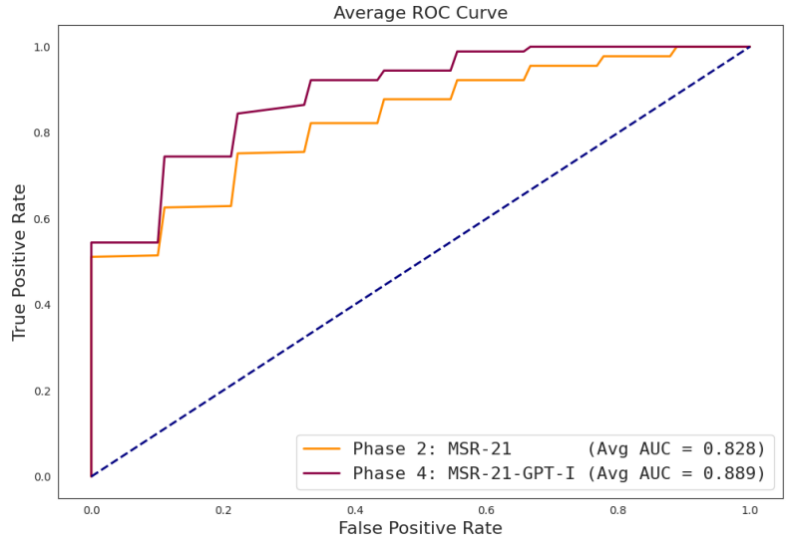

*Figure 10.* TNR comparison for MSR-21 and MSR-21-GPT-I datasets

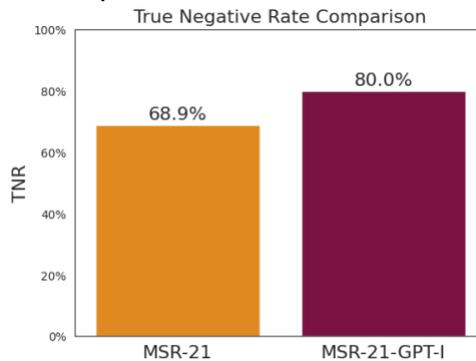

## 6. Discussion

### 6.1. Response to Research Questions

In this study, we investigated how students' academic writing profiles could be constructed and utilised in AV methods within educational contexts (RQ1). Additionally, we evaluated the profiles using datasets expanded with genAI-generated texts (RQ2) and texts generated by genAI explicitly tasked to mimic unique student writing styles (RQ3).

Our findings for RQ1 demonstrated that using lexical and syntactic features to create comprehensive student writing profiles resulted in strong AV performance across datasets, including the PAN-14 dataset, a diverse set of student essays (MGE-19), and a smaller dataset of technical software engineering reports (MSR-21). Notably, the profiles performed exceptionally well on the higher education datasets, validating their applicability for both general and technical academic writing tasks.

For RQ2, we observed that the developed writing profiles were robust in distinguishing GPT-authored texts from human-authored ones. This robustness highlights the potential of AV techniques to detect AI involvement in educational contexts, addressing key challenges in identifying AI-generated content discussed in the introduction. The comparative evaluation, while constrained by test set modifications, revealed that the expansions with GPT-generated texts achieved comparable or better results than the original datasets, particularly for software engineering reports. These findings demonstrate the method's reliability in detecting AI involvement without compromising accuracy when applied to diverse writing styles and formats.

In addressing RQ3, we evaluated the profiles against genAI explicitly instructed to mimic student writing styles. Our adapted FVD method performed significantly better on these impersonation expansions than on the original datasets, across all evaluation measures. This suggests that the profiles are not only robust to mimicry attempts but also capable of identifying subtle stylistic discrepancies introduced during human-AI collaboration. While GPT-4 was sometimes able to imitate vocabulary, typos, and answer organisation, it often exaggerated stylistic elements, such as using overly flowery language or unnatural turns of phrase, particularly for students with stronger vocabulary. These exaggerated features made the impersonations more distinguishable, highlighting the granularity and sensitivity of our AV approach to word and sentence-level stylistic features. Although it is unlikely that students would directly use genAI in this manner, these findings provide critical insights into how human-AI collaboration could be detected in adversarial scenarios. The simulation of challenging use cases where AI attempts to obfuscate authorship through mimicry allows our exploratory analysis to demonstrate the resilience of student writing profiles in capturing the dynamics of human-AI collaboration. This capability directly contributes to minimising gaps in existing AI detection tools, which often struggle with obfuscation and paraphrasing (Weber-Wulff et al., 2023).

This study's granular stylistic features, such as vocabulary and sentence structure, enhance explainable AI by providing interpretable insights into the differences between human and AI-generated texts. This transparency builds trust among educators and supports better understanding of human-AI collaboration while maintaining academic integrity. Unlike commercial detection tools, which often fail with obfuscated or edited texts (Weber-Wulff et al., 2023), our FVD method achieves high accuracy even in cross-topic and mimicry scenarios (Anderson et al., 2023). By analysing stylometric features at a granular level, our approach enables continuous authorship identification while also supporting students' academic development. Additionally, by accounting for linguistic diversity and by focusing on students' writing profiles (rather than on LLMs), our model reduces biases against non-native English speakers and unconventional writing styles, making it a fairer, more inclusive HAI collaboration solution.

Recognising that human-AI collaboration in writing is becoming an integral part of learning, our approach leverages authorship verification not as a tool for AI detection or punishment, but as a means to understand and support students' writing development in academic settings (Pedreschi et al., 2025). By tracking how writing evolves across drafts, our method helps identify whether and how AI assistance contributes to students' progress in areas such as critical thinking, coherence, and originality. While over-reliance on AI can lead to superficial or homogenised outputs (Fan et al., 2025), our approach promotes reflective and purposeful use by making the collaborative process more transparent and accountable.

### 6.2. Impact
Stylometric student writing profiles hold significant potential beyond AV, offering snapshots of students' writing capabilities and enabling longitudinal tracking of development across courses and disciplines. These profiles can enrich learning by providing personalised feedback and advanced analysis to educators and students. Educators could establish baseline profiles using supervised assignments and compare future submissions to monitor writing progress, ensure consistency, and identify potential academic integrity concerns. Students could receive actionable insights into writing aspects related to their vocabulary richness, sentence complexity, and writing patterns. These continuously monitored writing aspects can enable for targeted interventions such as clarity-focused exercises or vocabulary enrichment, led by educators. The maintenance and monitoring of academic writing profiles support tailored learning plans that can address individual challenges and foster academic growth.

**6.3. Limitations**

The small sizes of the datasets used for this study were fairly limiting for our research. While previous research (Hirst & Feiguina, 2007) demonstrates that combinations of POS and stylometric features perform reliably with limited training data, performance typically plateaus with larger datasets than those used here. Alternative approaches, such as augmenting datasets by generating permutations of known-unknown pairs (Potha & Stamatatos, 2014) could enhance future studies. Moreover, our study did not explore scenarios where humans and AI contribute within the same text, such as mixed sentences and/or paragraphs. While we believe our current model has the potential to quantify these contributions, this remains untested. Additionally, the tool has yet to be tested in live educational settings, where safeguarding student privacy and ensuring ethical data management are critical. To address these concerns, educational institutions must guarantee secure data storage and adherence to relevant privacy laws and institutional policies, similar to the protocols followed for LMS platforms and tools like Turnitin. Transparency is also critical—educators and institutions should clearly communicate to students how their data will be used. Above all, the deployment of such tools must prioritise and uphold the privacy and rights of students at every stage.

**6.4. Extensions**

Our methodology could be extended by replicating experiments on larger educational datasets, enabling control over confounding variables like text length, discipline, and age. Further optimisation could involve refining feature sets, applying techniques such as Recursive Feature Elimination (RFE) (Adamovic et al., 2019) or Principal Component Analysis (PCA) (Jamak et al., 2012), and analysing the impact of individual features (Potha & Stamatatos, 2014).

Future research could also prioritise high recall to minimise false negatives, ensuring students are correctly identified in educational settings. Adopting an education-specific evaluation framework and alternative metrics, such as $F_{1.5}$, could better address the need for fairness and accuracy in classifying authorship.

## 7. Conclusion

This study demonstrates the potential of stylometric academic writing profiles to advance AV within educational contexts, addressing critical challenges posed by human-AI collaboration. By integrating lexical and syntactic features into a transparent and robust AV framework, we effectively distinguished between student-authored and AI-generated texts, even under mimicry scenarios. This nuanced approach not only enhances academic integrity but also supports skill development by offering educators actionable insights into students' writing progress.

Our findings validate the applicability of writing profiles across diverse academic tasks and datasets, bridging gaps in existing detection tools that struggle with obfuscation or non-native writing styles. Importantly, the methodology fosters inclusivity and reduces biases, ensuring fair assessment practices. By providing interpretable and scalable solutions, this research underscores the transformative role of AV in education. Future directions include expanding dataset sizes, refining features, and exploring real-world classroom applications to further bolster transparency, ethical AI use, and personalised learning strategies.